\documentclass[letterpaper,conference]{ieeeconf}

\usepackage[pdftex]{graphicx}
\graphicspath{{./figures/}}

\usepackage{amsmath}
\usepackage{amsfonts}
\usepackage{amssymb}
\usepackage{amstext}
\usepackage{stmaryrd}

\usepackage{calc}
\let\labelindent\relax 
\usepackage{enumitem}

\usepackage{color}

\IEEEoverridecommandlockouts 
\usepackage{cite}

\usepackage{array}
\usepackage{booktabs}

\usepackage[caption=false,font=footnotesize]{subfig}

\usepackage{stfloats}

\begin{document}

\title{\huge Cheap or Robust? The Practical Realization\\ of Self-Driving Wheelchair Technology\vspace{-2mm}}

\author{
{Maya Burhanpurkar}$^{*}$,
{Mathieu Labb\'{e}}$^\dagger$,
{Xinyi Gong}$^{*}$,
{Charlie Guan}$^\ddagger$,
{Fran\c{c}ois Michaud}$^\dagger$, and
{Jonathan Kelly}$^{*}$
\thanks{This research was supported in part by funding from the National Sciences and Engineering Research Council of Canada (NSERC).\newline
\indent $^{*}$M.\ Burhanpurkar, Xinyi Gong, and J.\ Kelly are with the STARS Laboratory, Institute for Aerospace Studies, University of Toronto, Ontario, Canada, \texttt{\{maya.burhanpurkar,xinyi.gong,jonathan.kelly\}\newline@robotics.utias.utoronto.ca}.\newline
\indent $^\dagger$M.\ Labb\'{e} and F.\ Michaud are with the Interdisciplinary Institute for Technological Innovation (3IT), Universit{\'e} de Sherbrooke, Qu{\'e}bec, Canada, \texttt{\{mathieu.m.labbe,francois.michaud\}@usherbrooke.ca}.\newline
\indent $^\ddagger$C.\ Guan is with the Robust Robotics Group, Computer Science and Artificial Intelligence Laboratory, Massachusetts Institute of Technology, Massachusetts, USA, \texttt{cguan@mit.edu}.}
}

\maketitle
\vspace*{-6mm}

\begin{abstract}
To date, self-driving experimental wheelchair technologies have been either inexpensive or robust, but not both. Yet, in order to achieve real-world acceptance, both qualities are fundamentally essential. We present a unique approach to achieve inexpensive and robust autonomous and semi-autonomous assistive navigation for existing fielded wheelchairs, of which there are approximately 5 million units in Canada and United States alone. Our prototype wheelchair platform is capable of localization and mapping, as well as robust obstacle avoidance, using only a commodity RGB-D sensor and wheel odometry. As a specific example of the navigation capabilities, we focus on the single most common navigation problem: the traversal of narrow doorways in arbitrary environments. The software we have developed is generalizable to corridor following, desk docking, and other navigation tasks that are either extremely difficult or impossible for people with upper-body mobility impairments.
\end{abstract}

\section{Introduction}
\label{sec:introduction}

Electric power wheelchairs are often prescribed to individuals with mobility challenges. For many millions of users who suffer from a range of upper-body motor disabilities, such as quadriplegia, age-related hand tremors, Parkinson's disease, multiple sclerosis, and amyotrophic lateral sclerosis, it is impossible to operate an electric wheelchair using the standard joystick interface. These individuals must instead rely on other types of assistive control devices, such as sip-and-puff switches, which are typically extremely difficult to use. This results in degraded mobility, a lack of meaningful engagement with society, and ultimately reduced life expectancy \cite{keeler2010impact}. 

A robotic navigation system for electric wheelchairs, which would allow the chairs to self-navigate in home and workplace environments, would dramatically improve users' mobility. Currently, there are over 5 million power wheelchairs in use in Canada and the United States and an estimated 20 million such chairs deployed in G20 countries. Thus, the design of a cost-effective navigation system that can be retrofitted to these existing chairs is of great importance. Such a technology would substantially increase quality of life and societal engagement for this extremely vulnerable segment of society.


Past research efforts have focused on the development of self-driving wheelchair technologies using cost-prohibitive industrial sensors such as 3D laser scanners and advanced stereoscopic depth cameras, as well as high-performance computing hardware for data processing \cite{simpson2005smart}. This cost profile makes past research on self-driving wheelchairs infeasible for use in near-term consumer applications. Furthermore, previously-explored systems have lacked the requisite robustness for a consumer device.

\begin{figure}[t]
\vspace*{-2mm}
\centering
\setlength{\fboxsep}{0pt}%
\setlength{\fboxrule}{1pt}%
\fbox{\includegraphics[width=2.2in]{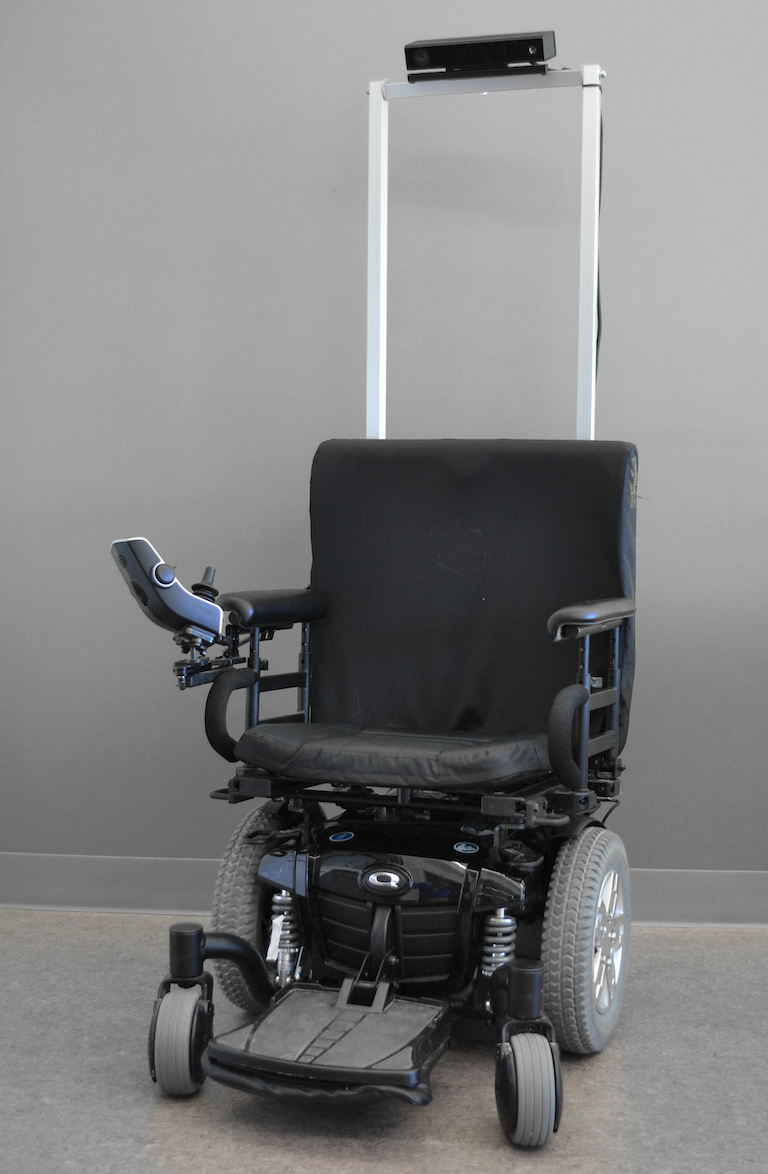}}
\caption{Autonomous wheelchair prototype, with the Microsoft Kinect 2 sensor mounted on an elevated frame to ensure an adequate field of view. The onboard navigation computer is mounted behind the seat.}
\label{fig:wheelchair}
\vspace*{-3mm}
\end{figure}

In this paper, we present a robotic navigation system for electric wheelchairs that provides a unique combination of attributes essential to a solution that is both economical and robust. By design, the system must be:

\begin{enumerate}[label=(\alph*)]

\item inexpensive --- uses only consumer grade-electronics and operates with reasonable computational demands;

\item portable --- able to be retrofitted to existing wheelchairs; 

\item robust and reliable --- does not rely on vision algorithms that are susceptible to poor lighting or difficulties with clutter, and uses specialized path planning to ensure ideal alignment for tasks such as doorway traversal;


\item rapidly deployable --- does not require an \textit{a priori} map, allowing for deployment in unfamiliar environments; 

\item non-proprietary --- uses open source software, leveraging existing open source libraries and components to reduce cost; and

\item readily extensible --- able to be extended to add new capabilities such as corridor following, desk docking, etc.
\end{enumerate}

Herein, we describe our efforts to address all of the above points above simultaneously, with the aim of developing a low-cost, practical system suitable for mass adoption and capable of robust performance under real-world operating conditions. Our prototype successfully achieves many of the design requirements while still meeting severe cost constraints, requiring less than \$2,000 US worth of components. As an example of the system's capabilities, we focus much of our discussion on the difficult task of doorway traversal.

\section{Related Work}
\label{sec:related_work}

A wide variety of assistive wheelchair navigation systems have been developed over the past 30 years, and we provide only a brief survey of related literature. Low-cost solutions are generally either limited and semi-autonomous \cite{miller1995design,shen2016lowcost}, require an external localization system \cite{garcia2013intelligent,kim2015robotic}, are able to operate over short distances only without global localization \cite{rockey2012low}, or lack global planning capabilities \cite{wu2013rgbd}. Fully autonomous navigation approaches typically rely on expensive laser scanner (lidar) sensors \cite{mandel2005towards,cruz2010slam,tang2012development,morales2013human,cavanini2014slam} (which do have the advantage of being able to operate outdoors \cite{gao2010towards, yokozuka2014development,schwesinger2017smart}). The goal of existing systems has usually been to provide a working solution in the target environment, without attempting to minimize cost.

The specific problem of doorway traversal, in particular, presents many challenges---no truly robust and cost-effective door detection and navigation system currently exists \cite{simpson2005smart}. Recent approaches utilize visual properties extracted from images, but such methods suffer from difficulties with lighting conditions as well as susceptibility to a variety of artifacts, leading to computationally demanding and unreliable solutions \cite{anguelov2004detecting}. For example, the three-camera visual approach in \cite{pasteau2016visual} relies on edge detection and vanishing point identification to extract trapezoidal (door-like) structures in the environment; similarly, the method in \cite{panzarella2016copilot} uses two cameras to identify occupied and free space. Neither approach is robust to poor lighting or visual clutter. Further, the planning mechanisms implemented in both cases may be insufficient for tight doors because the chair may not begin in a pose that is well aligned for traversal. In addition, the approach in \cite{panzarella2016copilot} lacks a definitive mechanism for navigating the common case of a corridor on the other side of the door.

One possible alternative to visual sensing is to employ active scanning devices. In \cite{cheein2010slam}, multiple 2D laser scanners are used to match input data to three potential doorway configurations. While successful in benign environments, 2D lasers are unable to detect objects above and below the plane of the laser, resulting in false positives. For example, desks and chairs may be detected as doors. Planar laser-based methods also are unable to verify that a detected doorway is traversable along the vertical axis.

In contrast to the above, the system we have developed makes use of a commodity RGB-D sensor to provide rich and reliable 3D spatial data, leverages custom and open source software for sophisticated data processing to ensure reliability, and considers unique characteristics of the hardware platform (the power wheelchair) during planning to enable successful operation in cases where other approaches fail.

\section{System Description}
\label{sec:systemdescription}

Our prototype navigation system is based on a standard commercial electric wheelchair, shown in Figure \ref{fig:wheelchair}, to which we have retrofitted a Microsoft Kinect 2 sensor, wheel encoders, and an onboard computer (Intel i7 processor, 4 cores). The total cost of the additional hardware is less than \$2,000 US (retail), and would be even lower in an OEM production scenario. While previous research has focused on varying aspects of autonomy, including wall following, obstacle avoidance, and doorway traversal, modern simultaneous localization and mapping (SLAM) software enables the unification of these functions within a common navigation framework.


\begin{figure*}
\begin{center}
\includegraphics[width=5.5in]{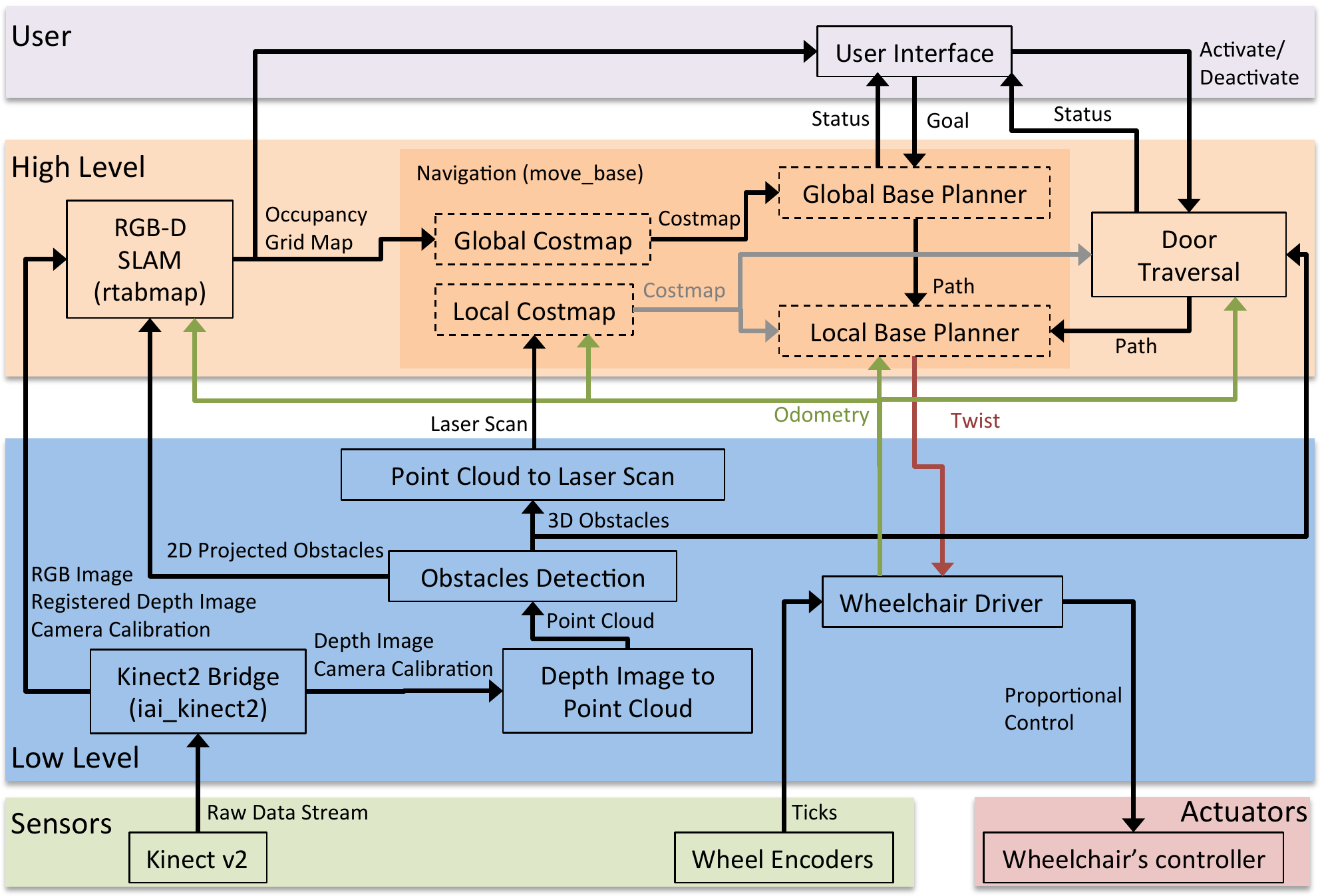}
\caption{System architecture, designed in four layers: \emph{sensing and actuation}, \emph{low-level processing}, \emph{high-level processing}, and \emph{user interaction}. The individual software components are built around the Robot Operating System (ROS), which facilitates easy integration and allows us to leverage open source libraries.}
\label{fig:system}
\vspace*{-5mm}
\end{center}
\end{figure*}

Figure \ref{fig:system} illustrates the system architecture. We make significant use of the Robot Operating System (ROS) \cite{quigley2009ros} to provide underlying support for navigation-related tasks. However, we stress that ROS alone provides only a limited portion of the software used to drive the chair---the majority of the code is bespoke and carefully tuned for our application. The navigation system is divided into four layers: 1) a sensor and actuator (hardware) layer (green and red); 2) a low-level processing layer (blue) which filters and translates raw sensor and actuator data streams into more useful formats; 3) a high-level processing layer (orange) containing three submodules: SLAM, navigation, and door traversal; and 4) a user interface layer (purple) that provides visualization and navigation control.


In a typical deployment scenario, an initial map of the environment is built by an operator who manually guides the wheelchair to visit all of the locations where the platform will be expected to navigate. During the mapping process, the SLAM module generates a 2D floor plan, while automatically detecting when the wheelchair revisits a previously-traversed area (to maintain map consistency). The floor plan is then validated by the operator and corrected, if necessary, using an interface tool currently in development. Once the map is available, a wheelchair user can click on a destination location, and the navigation software will compute a collision-free trajectory to reach the goal. While navigating, any dynamic obstacles along the path are automatically avoided. If the wheelchair is shut down and restarted, the SLAM module can re-localize anywhere in the existing map, without the need to start from the same initial location. 

The door traversal module is a recent addition to the navigation system, and is able to operate independently of a full environment map. We have found that robust door traversal requires specialized processing beyond that of the baseline SLAM solution. Our doorway traversal software is capable of reliably negotiating doorways and other narrow passages while following a smooth and predictable path to the destination location. Testing has shown that the system operates well in complex environments with diverse geometries and scales.

The following subsections describe the low-level processing layer and the three key submodules of the high-level layer.

\subsection{Low-Level Processing Layer}

Low-level processing is performed by five software components: the Kinect 2 Bridge, the Depth Image to Point Cloud module, the Obstacle Detection module, the Point Cloud to Laser Scan module and the Wheelchair Driver module.

Kinect data is acquired using the \texttt{iai\_kinect2} ROS package \cite{iai_kinect2}. The maximum frame rate is set to 10 Hz to limit CPU usage. However, the module still requires 100\% of one CPU core to decode the USB 3 data stream and to register the depth images. Registered depth images are required by the RGB-D SLAM module so that loop closure transformations can be correctly computed.

Depth images from the Kinect are projected into 3D space by the Depth Image to Point Cloud module, generating a point cloud for obstacle detection. The depth image has a resolution of 512 $\times$ 424 pixels, which results in a cloud of 217,088 points. To reduce CPU usage, the depth image is decimated by a factor of four (to 256 $\times$ 212 pixels), so that the resulting cloud has a maximum of 54,272 points. Further reductions in size are achieved using a voxel filter with a size of 5 cm \cite{rusu2011pcl}. For navigation in most environments, a 5 cm voxel size represents a good tradeoff between computational load and sufficient precision for obstacle avoidance (based on the wheelchair dimensions and minimal clearance requirements).


Filtered point clouds are used by the Obstacle Detection module to segment the ground plane and identify potential navigation hazards. A normal vector is computed for each point, to determine if the point lies on a plane parallel to the ground---any vector that lies more than 20$^{\circ}$ away from the `upward' axis is considered to belong to an obstacle. The 20$^{\circ}$ threshold was selected to be large enough to be robust to Kinect's noise and small enough to avoid labelling real obstacles as part of the ground. Remaining planar surfaces with centroids at a height of less than 10 cm are labelled as `ground'. Points belonging to obstacles are projected onto the ground plane for use in 2D mapping. 

Movement commands are sent to the wheelchair through the Wheelchair Driver module, which translates velocity messages (twists) from the Navigation module into the format required by the wheelchair's onboard controller (in a PID loop). The module also computes wheel odometry information using knowledge of the wheel sizes. This odometry data is then published to other modules and is used as feedback for the PID controller. The onboard computer is connected to the wheelchair's existing (factory) control interface, emulating the standard wheelchair joystick. The control loop runs at 10 Hz, matching the maximum rate the wheelchair's existing controller can sustain; the proprietary internal controller of the wheelchair cannot be configured to run faster than 10 Hz. Increasing this control rate would help to enable even more accurate path following.

\subsection{RGB-D SLAM}

The RGB-D SLAM module provides autonomous mapping capabilities for the wheelchair. Based on the open source RTAB-Map ROS package \cite{labbe14online}, the module creates a detailed 2D occupancy grid map from RGB-D images. The map is a graph, where each node stores a synchronized RGB image, a registered depth image, a set of 2D projected obstacles, and odometry information. Using the poses in the graph, a global occupancy grid can be generated by assembling all of the local occupancy grids. The complete map is updated at 1 Hz to limit computational load. At each map update, a new RGB image is visually compared with past images in the map to identify loop closures, based on a bags-of-words approach \cite{labbe13appearance}. When a loop closure is found, any accumulated drift in the odometry is corrected. The local occupancy grid maps are then re-assembled with the corrected wheelchair poses.

The module can be configured in one of two modes: SLAM or localization. The SLAM mode is used to create the initial map of the environment. After creating the map, localization can be activated, without continuously adding new data to the map. This limits computational and memory requirements. Note that it is possible to switch back to SLAM mode to remap an area that has changed significantly or to extend the current map to new locations.


Although incremental motion could be computed using visual odometry (VO), we have found that, with a single camera setup, VO is insufficiently robust for our application. Because of the limited field of view of the camera and because the wheelchair sometimes has to navigate in areas with insufficient visual features, VO can become `lost' easily. We utilize wheel odometry instead, even if it is slightly worse than VO in environments that are visually rich. To reduce odometric drift, 2D projected obstacles from consecutive graph (map) nodes are used to refine the pose-to-pose transformations. The 2D projected obstacle data is also employed to refine any identified loop closures.

\subsection{Navigation}

Our Navigation module is based on the \texttt{move\_base} ROS package \cite{officeMarathon}. Given the wheelchair's configuration (i.e., geometry, differential drive, maximum acceleration, and speed), the Local Base Planner module can provide appropriate velocity commands to reach a goal location while avoiding obstacles in the local cost map. The cost map is a 2D occupancy grid where obstacles are inflated by a fixed radius (e.g., generally the robot radius) so that the planner can determine how far away it should pass to safely avoid collisions.

The local cost map has a fixed size of 4 m $\times$ 4 m (the robot is always at the centre) and is updated with the latest sensor readings. To handle dynamic environments, the local cost map is `cleared' of obstacles at each update if possible (e.g., if an obstacle has disappeared). To do so, 2D ray tracing (when using a 2D sensor like a 2D laser scanner) or 3D ray tracing (when a 3D point cloud is used) is performed. The latter technique is the most computationally intensive, and so for efficiency we use the first approach, relying on the \texttt{pointcloud\_to\_laserscan} ROS package \cite{pointcloudToLaserscan} to carry out the 3D to 2D conversion and ray tracing.

For fully autonomous navigation, the Global Base Planner module of \texttt{move\_base} is utilized \cite{brock1999high}. Given the global occupancy grid created by RTAB-Map, the current pose, and the desired goal, a path is planned through the empty cells in the global map. If the goal coincides with an obstacle, or if the location cannot be reached (e.g., a new obstacle is blocking the way), the planning step fails and the user is notified. When a complete path is able to be computed, it is sent to the Local Base Planner module and a series of sub-goals are selected in incremental local cost maps. The Local Base Planner is thus not constrained to exactly follow the global plan,  allowing for maneuvering around dynamic obstacles. When RTAB-Map corrects the global map or re-localizes, the global path is recomputed accordingly from the current map location.

\vspace{-1mm}
\subsection{Door Detection and Traversal}
\label{sec:doors}

In general, previous attempts at developing an assistive wheelchair with door traversal capabilities have failed due to sensor limitations or difficulties in planning for correct wheelchair-to-door alignment. Our approach utilizes 3D depth data to identify doorways by exploiting their planar nature. The method is motivated by two principles: 1) doorways must contain free space, and 2) the walls that support a door are almost always planar, or contain planar regions even if cluttered with furniture or other objects. Using these constraints, we apply a subsampling and clustering approach, which leads to robust traversal. This is similar to the detection method described in \cite{derry2013automated}, although we consider a wider variety of possible door configurations.

Conceptually, the algorithm first checks the surroundings for walls, then searches for free space within or between them, generating potential doorway candidates. If a gap of sufficient width is detected between the walls and free space is detected along the plane of the door, a doorway is validated. The detection algorithm provides the doorway edges to the door traversal planning routine. Using the normal vector to the plane of the door, the door traversal routine sends a series of global goals to the ROS navigation stack, perpendicular to the plane of the door to ensure alignment for successful traversal. The process can be segmented into four steps: filtering and wall extraction, door candidate identification, validation, and traversal. The first three steps are repeated until a door is successfully found or until the number of points remaining in the input point cloud falls below a threshold value. 

\textit{Filtering and Wall Extraction}: The input to the algorithm is the Kinect 3D point cloud, processed by the Point Cloud Library \cite{rusu2011pcl}. The plane of the most prominent wall remaining in the input cloud parallel to the $z$-axis (vertical) is then identified using a Random Sample Consensus (RANSAC) estimator \cite{fischler1981random}. In order to limit the influence of obstacles near the ground, the wall plane cloud is filtered in $z$, such that only the midsection of the wall remains.

\textit{Door Candidate Identification}: Filtered wall planes are stored in a tree and clustered using a Euclidean cluster extraction routine. These clusters define potential wall segments surrounding the door. Each valid cluster is added to an array of candidate clusters. Typically, one or two clusters are added per iteration of the algorithm–--two in the case of a door defined by a gap in a single plane, and one in the case of a door contained between two planes. During each iteration of the door detection algorithm, every previously untested combination of two clusters is checked as a potential door candidate. If the candidate planes are parallel to within a pre-defined tolerance, they are passed to a Single Plane Detection routine; if they intersect, they are passed to a Double Plane Detection routine. In this way, any configuration of door can be detected.

\begin{enumerate}[label=(\alph*)]
\item Single Plane Detection: This routine identifies doorways bounded by a single plane, such as hallway entrances and doorways with non-protruding hinged doors. First, the direction of chair tilt is determined. Next, the rightmost cluster is identified by comparing the minimum and maximum $y$-values for points in each cluster. From this information, the left and right edges of the door are extracted from the cluster clouds through a minimum-maximum search within each cluster. The left and right door coordinates are returned.

\item Double Plane Detection: This routine identifies doorways bounded by two planes, such as protruding hinged doors that occlude a section of the wall surrounding a doorframe. The intersection of the planes at the height $z$ = 0 is found, and used in a nearest neighbours search over the points in each plane to determine the left and right door edges.
\end{enumerate}

\textit{Validation}: The door edges are passed to a validation routine that checks whether the resulting door width is acceptable and whether there is sufficient free space for the chair to pass. If the number of `stray' points between the door planes is below a threshold value (non-zero due to sensor noise) the doorway candidate is validated.

\textit{Traversal}: Using the left and right door edges, a series of goals up to, including, and exceeding the centre of the door frame, is sent to the Local Base Planner. This is done to ensure proper alignment with the door (tests conducted by sending only one goal at the midpoint of the left and right door edges often failed in low clearance doors, whereas sending multiple intermediate goals results in robust performance). The number of goals sent varies depending on the starting distance from the door. After each goal is sent and reached, the input obstacle cloud is filtered along the $x$-axis by the distance the chair has moved in the $x$ direction, to prevent obstacles or walls on the opposing side of the doorway from being detected, as the chair continuously re-evaluates the position of the door.

Obstacles near a door, such as free standing coat racks or water fountains, do not confound the algorithm since a) an object separated from the wall plane in the $x$-axis will likely not be included in the plane identified by RANSAC, b) in the event that it is, the obstacle likely will not contain enough points for it to pass clustering validation, and c) if it is accepted as a cluster, it will be rejected by the validation routine. Finally, if Door Candidate Identification or Validation fails, the most recent wall being examined is removed from the 3D point cloud and a new wall is searched for with the remaining points.

\section{Experimental Results}
\label{sec:results}

We have carried out a comprehensive series of experiments to evaluate and validate all aspects of the wheelchair platform. This includes a major simulation campaign, made possible by fully modelling the chair in the Gazebo simulator \cite{koenig04design}, and several studies at the Universit{\'e} de Sherbrooke and the University of Toronto. Below, we report on our results to date, which involve unmanned testing to ensure all safety concerns are addressed; in Section \ref{sec:discussion} we review a number of the challenges that remain, and in Section \ref{sec:conclusions} we describe plans for upcoming trials with potential end users.

\vspace{-1mm}
\subsection{Simultaneous Localization and Mapping}
\label{sec:results_slam}

\begin{figure}
\centering
\setlength{\fboxsep}{0pt}%
\setlength{\fboxrule}{1pt}%
\fbox{\includegraphics[width={\columnwidth-4pt}]{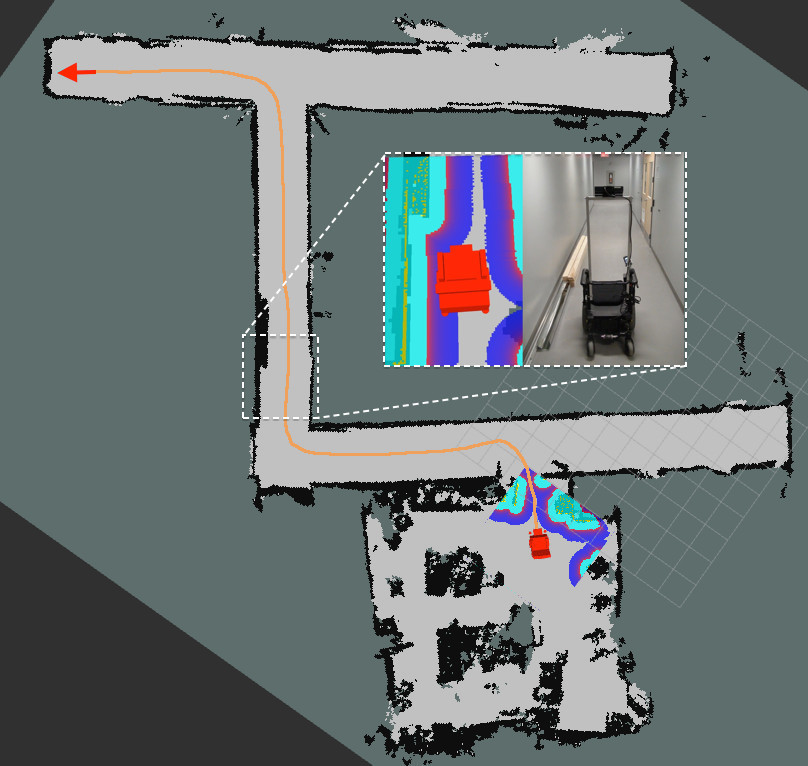}}
\caption{Map created using SLAM. The orange line is an example of a planned path from the bottom of the map to the goal, represented by the red arrow at the top left (pose with desired orientation). The close-up view shows how the wooden object on the ground is detected. The cost map is represented in colour: yellow pixels are obstacles and cyan represents regions where the planner cannot plan a path because of the risk of collisions.}
\label{fig:global_plan}
\end{figure}

The SLAM module, based on RTAB-Map, has an extensive heritage (e.g., winning the IROS 2014 Microsoft Kinect Challenge), and has been shown to operate robustly in a wide range of environments. As an example, Figure \ref{fig:global_plan} shows the map produced after a standard mapping run at the Universit{\'e} de Sherbrooke. Light grey pixels are empty space and black pixels are obstacles. Since the wheelchair has a front-facing camera only, it cannot see behind itself and thus cannot localize when navigating the same corridor in reverse direction. This problem sometimes creates a double-wall effect when the odometry drifts significantly. During the mapping phase, we usually carry out a series of 360$^{\circ}$ rotations so that loop closures can be detected at locations seen from the opposite direction. To produce the map shown, full rotations were carried out at the corners of the corridors. The zoomed-in section of the map shows that the wooden object lying on the floor is correctly detected during mapping (the wall is thicker and the local cost map is suitably inflated).

\vspace{-1mm}
\subsection{Autonomous Navigation}

\begin{figure}
\centering
\setlength{\fboxsep}{0pt}%
\setlength{\fboxrule}{1pt}%
\fbox{\includegraphics[width={\columnwidth-4pt}]{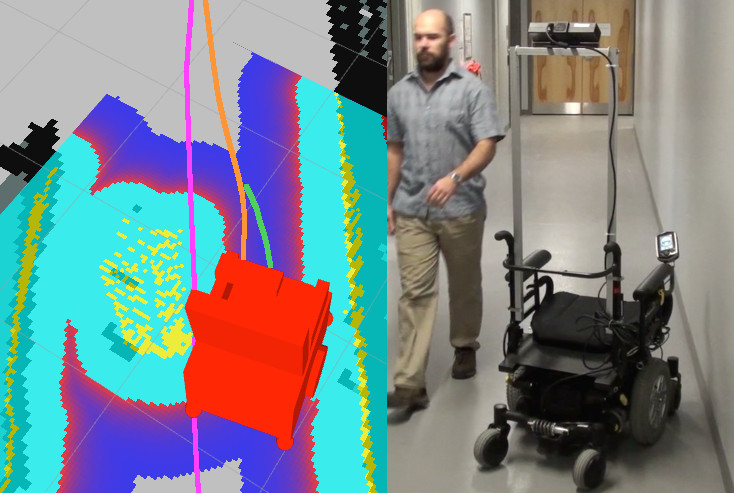}}
\caption{The wheelchair autonomously avoids a person walking down the corridor. The purple line is the original planned path; the orange is the new global plan, updated with the new obstacle detected. The green line is the actual velocity command sent by the Local Base Planner to the chair controller.}
\label{fig:avoid}
\vspace*{-3mm}
\end{figure}

The full navigation system has been tested through more than 10 km of autonomous driving, in varying situations and with an assortment of static and dynamic obstacles. This testing has enabled us to refine the system, such that it now operates very reliably in the vast majority of cases. An example of dynamic obstacle detection and avoidance is shown in Figure \ref{fig:avoid}. Referring to Figure \ref{fig:global_plan}, the orange line represents the path planned by the Global Base Planner using the available global map. A user-specified goal is shown at the top left (red arrow). This initial path is sent to Local Base Planner to be executed. Figure \ref{fig:avoid} shows the system reacting to a (previously unseen) person walking directly on top of the original path (purple), obstructing progress. The cost map is automatically updated with the dynamic obstacle and a new plan is formulated (orange). The green line represents the actual velocity commands sent by Local Base Planner to the wheelchair controller.

\subsection{Doorway Traversal}
\label{subsec:door_testing}

Since door traversal is a critical capability for an practical autonomous or semi-autonomous assistive wheelchair platform, substantial testing of the traversal algorithm was first carried out in simulation (where a wide variety of doorways could be evaluated). Based on these results, we conducted more than 100 tests with real-world doors in buildings at the University of Toronto.

\begin{figure*}
\centering
\includegraphics[width=6in]{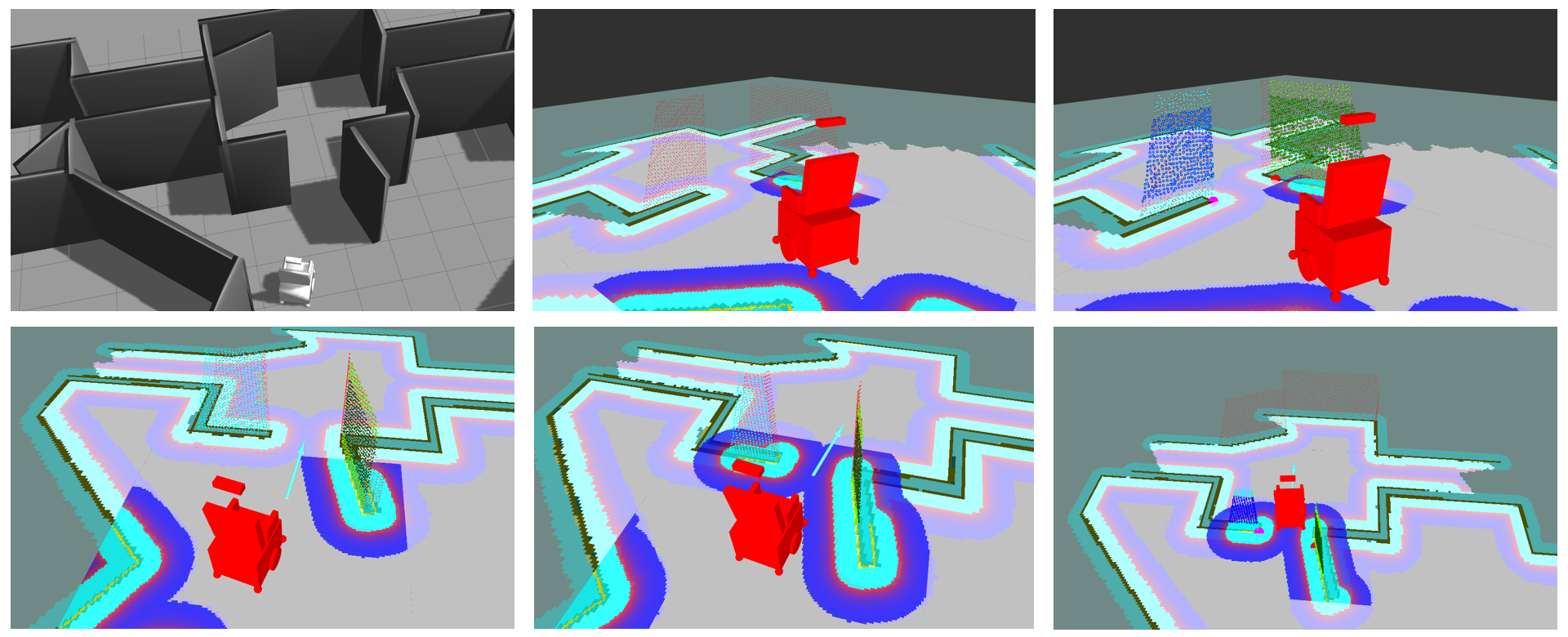}
\caption{A door traversal simulation experiment. Top row (left to right): simulation environment in Gazebo; the obstacle cloud in RViz, and the cloud processed by the door detection algorithm (light blue = plane 1, dark blue = plane 1 midsection, light green = plane 2, dark green = plane 2 midsection, pink and red circles = left and right edges of the door respectively determined using the point of intersection of plane 1 and plane 2 at $z$ = 0). Bottom row: The chair traverses the door in three steps: first aligning itself with the door, then traveling to the middle of the door frame, and finally traveling an additional distance forward from the frame.}
\label{fig:door_traversal}
\vspace*{-4mm}
\end{figure*}

In simulation, a map with 20 unique door configurations, allowing for 40 unique traversal situations, was constructed. Figure \ref{fig:door_traversal} shows an example of one simulated environment. Five tests were conducted on each side of each door with a 100\% success rate for all 200 tests. Additionally, several maps were constructed that included obstacles (such as bookshelves and coffee tables) near walls surrounding the doors, to ensure the effectiveness of the algorithm even in the presence of clutter. In these cases, door detection and traversal once again performed with a 100\% success rate.

\begin{figure*}
\centering
\includegraphics[width=0.67\columnwidth]{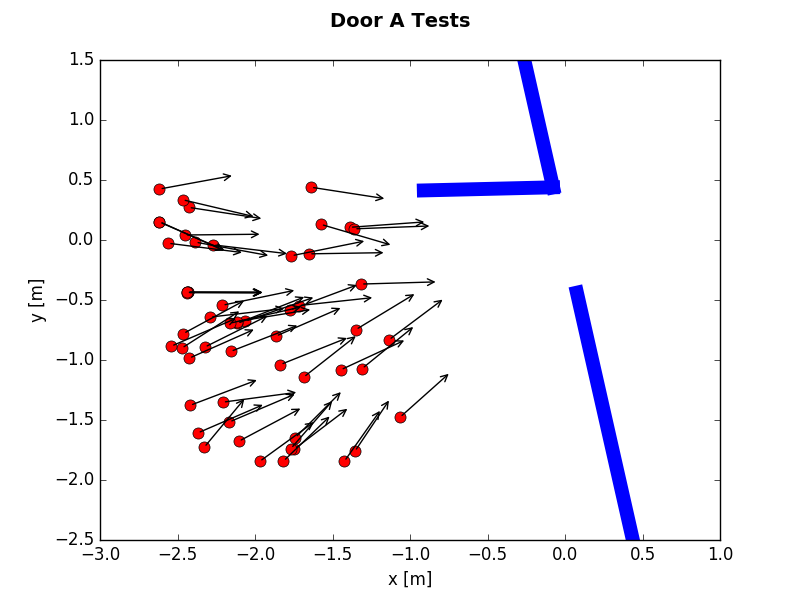}
\includegraphics[width=0.67\columnwidth]{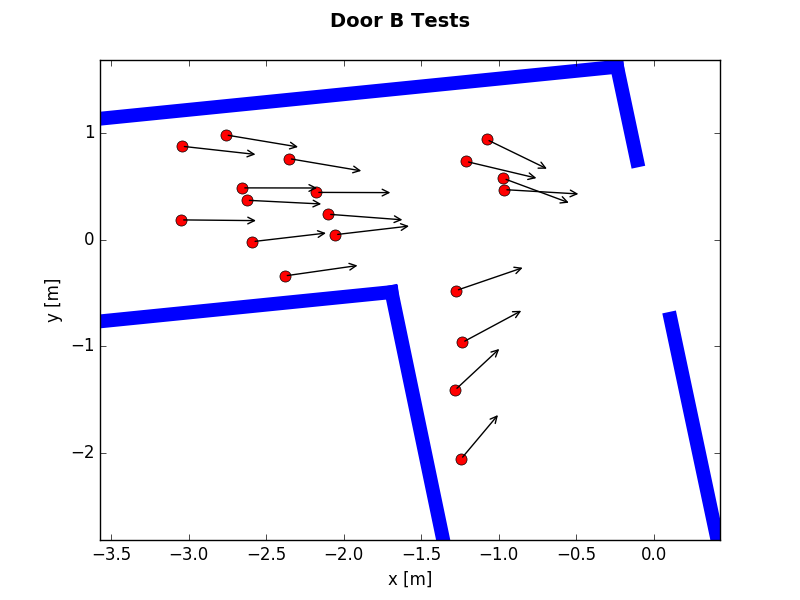}
\includegraphics[width=0.67\columnwidth]{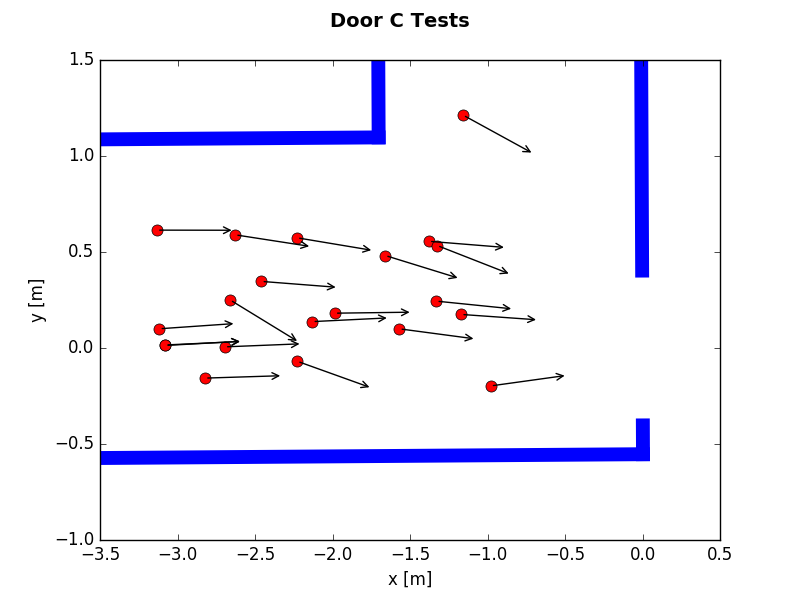}
\caption{Doorway traversal tests carried out at the University of Toronto. Each plot shows a top-down view of the door configuration (see Section \ref{subsec:door_testing}). Planar surfaces are shown in blue. Each red dot specifies the position of the centre of the wheelchair at the start of an experiment, while the corresponding arrow indicates the initial orientation of the chair at the start of the traverse.}
\label{fig:door_testing}
\vspace*{-1mm}
\end{figure*}

At the University of Toronto, the chair was tested on 8 different door configurations dozens of times each, again with a 100\% success rate (after initial failures that led to revisions to the algorithm). Formal tests were conducted on three representative doors, as shown in Table 1 and Figure \ref{fig:door_testing}. Door A was a doorway with a hinged door protruding, testing the Double Plane Detection method. Door B was a double doorway, testing the ability to detect large openings. Door C was a doorway without a hinged door, testing the Single Plane Detection method. Tests were conducted at a range of angles and distances from the door; plots of the initial wheelchair locations and orientations for each door are shown in Figure \ref{fig:door_testing}. The chair was able to navigate effectively throughout the entirety of the range of situations in which it was tested, including cases where only one plane could be detected.

\begin{table}[t]
\renewcommand{\arraystretch}{1.1}
\caption{Doorway Traversal Experimental Results}
\label{tab:doors}
\centering
\begin{tabular}{>{\raggedright}p{3.3cm} c  c  c}
	\toprule
	\textbf{Evaluation Metric} & \textbf{Door A} & \textbf{Door B} & \textbf{Door C} \\
    \midrule
	Number of Tests & 56 & 19 & 21 \\ \midrule
    Detection Success Rate & 100\% & 100\% & 100\% \\ \midrule
    Traversal Success Rate & 100\% & 100\% & 100\% \\ \midrule
    Mean Perceived Door Width & 0.89 m & 1.50 m & 0.89 m \\ \midrule
    True Door Width & 0.89 m & 1.51 m & 0.89 m \\ \midrule
    Door Width Std.\ Dev. & 0.05 m & 0.04 m & 0.03 m \\ \midrule
    Max.\ Distance Tested & 2.91 m & 3.17 m & 3.20 m \\ \midrule
    Min.\ Distance Tested & 1.37 m & 1.08 m & 1.00 m \\
    \bottomrule
\end{tabular}
\vspace*{-1mm}
\end{table}

\section{Discussion and Ongoing Challenges}
\label{sec:discussion}

While we have developed an initial prototype system that performs well in the majority of situations, a series of failure modes and corner cases remain to be investigated. For instance, as with any sensor, the Kinect 2 has some critical limitations. In particular, the unit can have difficulty registering accurate depth information in certain environments. Highly reflective surfaces may cause the sensor to return false depth data, while light-absorbent materials often produce a very low return. Transparent and translucent objects also cause erratic performance. Further, the Kinect 2 is unable to operate outdoors in bright sunlight, due to saturation of the IR receiver. Many of these issues could be mitigated by augmenting the system with other sensor types, although costs would increase.

For door traversal specifically, tuning was required initially to ensure reliable performance. Doors with reflective surfaces (even lower metal `bumper' strips) sometimes made proper plane extraction difficult. Door traversal also relies on stable incremental motion estimates. In cases where VO failed, the doors would successfully be traversed, but this would sometimes require the chair to rotate in place, which is undesirable in real-world usage. In addition, because of the sensor noise, the cost map inflation radius  must be relatively high to ensure safety, resulting in a very narrow section of the map in which goals can be sent through the door frame. This is, in part, due to downsampling of the point cloud data and the fact that most doors only provide a few centimetres of clearance on either side of the chair. The result is a need to reevaluate incremental planner goals, searching for nearest neighbour points that do not lie within the inflated safety radius. The heuristic could be relaxed if the point cloud were not downsampled, but processing requirements would increase. 




At present, our system assumes that the wheelchair is navigating on a planar floor surface (i.e., in 2D), without any ramps or elevators. It would be possible with the current framework to support a 3D surface by using 6-DOF odometry instead of the 3-DOF odometry currently employed (i.e., wheel odometry). More precise odometry would be also needed, as the chair would be able to move in a larger configuration space.

As mentioned in Section \ref{sec:results_slam}, having a rear-facing camera would assist in localization when traversing the same area in the reverse direction. No 360${^\circ}$ rotations would be required to correctly optimize the map, although it is always desirable to have many different views of the same environment for better localization. We note, however, that our computing resources are already used at nearly maximum capacity (e.g., processing the data stream from one Kinect 2 uses 100\% of one core of the CPU).

The frequency and latency of the sensing and control loops necessitate limiting the wheelchair velocity to a modest walking pace, to ensure sufficient time to respond to dynamic obstacles. The system is also dependent on the latency of the proprietary wheelchair controller, which we do not have direct access to. For example, even when issuing commands directly with the standard wheelchair joystick, delays between 100 ms to 400 ms are observed. The wheelchair also has caster wheels that create friction. If the caster wheels are perpendicular to the frame of the chair and a forward velocity command is sent, the wheelchair will rotate slightly until the casters are aligned with the desired travel direction. When navigating around nearby objects at a low speed, commands sent by the Local Base Planner are poorly executed, making difficult to maintain precise control under such conditions.


\section{Conclusions and Future Work}
\label{sec:conclusions}

This paper presented a cost-effective and robust autonomous navigation system for existing power wheelchairs. Based on an inexpensive sensor suite (an RGB-D sensor and wheel odometry), the various modules of the system (SLAM, navigation, and door traversal) function synergistically to enable reliable operation under real-world conditions. Our goal has been to develop a system that satisfies the characteristics listed in Section \ref{sec:introduction}. We continue to work to address various confounds and corner cases, with rigorous testing and validation. Full navigational autonomy has the potential to improve the safety of users and those around them, while greatly reducing operator fatigue.

We have now begun testing of our development platforms in busy home, office, and retail environments, in order to assess performance with end users. This testing is being carried out under the guidance of trained occupational therapists, ensuring that we meet the needs of the target community.

\section*{Acknowledgment}

This work was supported in part by NSERC Canada and by Cyberworks Robotics, Inc. The authors wish to thank Rodolphe Perrin and Daniel Elbirt for their assistance with data collection.


\bibliographystyle{IEEEtran}
\bibliography{references.bib}

\end{document}